# Agentic Nesting: A New Methodology for Existing Enterprise Application Integration and Services

**Xi Wang**[*1,2] **Kun Li**[2] **Xianyao Ling**[2] **Gang Yin**[2] **Liang Zhang**[1,3]
**Jiang Wu**[3] **Wenbo Lei**[4] **Jun Xu**[5] **Annie Wang**[6]
**Fu Zhang**[7] **Weizhe Wang**[1,8]

[1]Tsinghua University [2] Cross-strait Tsinghua Research Institute
[3]OceanBlue Construction Co. Beijing, Ltd. [4]Beijing Yixin Technology Co., Ltd.
[5]Hunan Jiace Evaluation Information Technology Service Co., Ltd. [6]Kalavai Corp
[7]Huacai Technology (Beijing) Co., Ltd. [8]Beijing Eastern Golden Info Technology Co., Ltd.

xi-wang19@mails.tsinghua.edu.cn likun@ctri.org.cn xianyao.ling@ctri.org.cn
yingang@ctri-ai.cn liang-zh19@mails.tsinghua.edu.cn jiang.wu@rcytgs.com
lwb@it-hdz.com xujun@jiacetest.com annie@kalavai.net
zhangfu@huacaizhaoyu.com wangwz20@mails.tsinghua.edu.cn

**Abstract:** Enterprise operations extensively rely on multiple heterogeneous business systems and information applications, which also result in severe data silos and process fragmentation. Enterprises have invested considerable financial and material resources in building these applications, however, effectively leveraging and orchestrating them remains a formidable challenge. Conventional approaches to enterprise application integration, encompassing middleware architectures such as Enterprise Service Bus (ESB), API gateway infrastructures, and Robotic Process Automation (RPA), suffer from inherent limitations like high architectural coupling, escalating operation and maintenance costs, and limited intelligence capabilities. This paper proposes Agentic Nesting, a multi-agent collaboration framework in which existing enterprise applications are encapsulated as autonomous AI agents within a hierarchically nested structure. Rather than flat interconnection, agents are organized into layered stewardship topologies that mirror the compositional complexity of enterprise ecosystems. The framework extracts a digital agent proxy from each legacy application to enable natural-language interaction and autonomous manipulation, coordinates multiple agents through a central orchestrator for task decomposition and dynamic dispatching, and exposes a unified conversational interface for cross-application querying and process orchestration. The main contributions of this paper are the proposition of the "Application-as-Agent" integration paradigm and the "Conversation-as-Integration" interaction philosophy, together with an exploration of the generalization potential of this methodology in scenarios encompassing heterogeneous system coordination, and large-scale data applications.

## 1. Introduction

### 1.1 Background

Within the contemporary landscape of enterprise digitalization, a representative organization typically sustains a substantial portfolio of business systems, ranging from several tens to multiple hundreds. These systems span diverse categories including Enterprise Resource Planning (ERP), Customer Relationship Management (CRM), Office Automation (OA), and Database Management Systems (DBMS), Data Warehouse (DW), various industry-specific applications, data analytics tools, and instant messaging systems. These systems are typically developed independently by different vendors during different periods, employing heterogeneous technical architectures, data models, and communication protocols; at the database level, diverse query languages, schema structures, and access interfaces further compound this heterogeneity, thereby giving rise to the pronounced phenomenon of data silos [1]. Employees must frequently switch among multiple systems during daily operations to perform manual data migration, information reconciliation, and report consolidation, which not only degrades operational efficiency but also introduces risks of human error.

Enterprise Application Integration (EAI) has historically represented a focal area of sustained scholarly inquiry within the information technology discipline. From early point-to-point integration patterns, to centralized architectures based on the Enterprise Service Bus (ESB), and to more recent developments such as Integration Platform as a Service (iPaaS) and Robotic Process Automation (RPA), each generation of technological solutions has mitigated interoperability challenges within its specific historical context. However, these solutions are fundamentally interface-oriented integration paradigms, which are predicated upon the prior definition of data mapping rules, message format specifications, and invocation protocol standards among systems. Even in the domain of database integration, traditional solutions likewise rely upon ETL pipelines or federated query middleware, requiring rigorous pre-configuration of schema mappings and access permissions. When business requirements evolve, the costs of modification and maintenance escalate significantly. More critically, such paradigms are ill-equipped to handle unstructured, highly ambiguous complex tasks and lack comprehension of high-level business semantics and adaptive capabilities.

* Corresponding author.

## 1.2 Agentic AI Development

The breakthrough advances in Large Language Models (LLMs) [2][3], rooted in the Transformer architecture's capacity for modeling long-range dependencies [4] and benefiting from the emergent phenomena revealed by Scaling Laws [5], have opened novel pathways for addressing the aforementioned predicaments. Based on the emergent capabilities of in-context learning, reasoning, and code generation arising from massive corpus pre-training [6][7], LLM-based AI agents possess core capabilities including natural language understanding, reasoning and planning, tool invocation, and code generation, enabling them to comprehend high-level human intentions and autonomously execute complex tasks.

During the recent triennial period, both scholarly and industrial sectors have observed an exponential expansion in the development of autonomous agent frameworks, including single-agent architectures such as ReAct [8] and Tool-Use Agent [9], as well as multi-agent collaborative systems exemplified by MetaGPT [10], AutoGen [11], CrewAI [12], continuously expanding the capability boundaries of AI agents. ReAct [8] proposes the paradigm of interleaving reasoning and acting, enabling large language models to dynamically invoke external tools during chain-of-thought reasoning processes [13], thereby realizing autonomous decision-making that thinks and acts concurrently. Related surveys have systematically reviewed research advances in autonomous agents based on large language models across dimensions including perception, planning, memory, and tool usage [14], whereas comprehensive reviews within the multi-agent systems research community have concentrated on cooperative mechanisms, inter-agent communication protocols, and collective intelligence emergence among agents [15].

Concurrently, agent communication protocols are rapidly evolving. The Model Context Protocol (MCP) proposed by Anthropic [16] provides standardized external tool access interfaces for models. Google's Agent-to-Agent (A2A) protocol [17] endeavors to establish standardized specifications for agent discovery and cross-platform interoperability; while the Agent Communication Protocol (ACP) extends this standardization effort toward more universally applicable agent communication specifications. Survey research on large language models provides a systematic perspective for understanding the evolution of agent technologies [18]. Studies such as Gorilla demonstrate that large language models can connect to massive APIs through retrieval-augmented techniques to expand capability boundaries [19]. Surveys in the tool learning domain further classify tool usage into four hierarchical levels: perception, learning, invocation, and creation [20]. Besides, RestGPT explores technical pathways for connecting LLMs with real-world RESTful APIs [21]. Nevertheless, current mainstream practice still models external systems as tools-agents access system APIs via function calls to retrieve data or execute operations. Although this "Application-as-Tool" paradigm offers simplicity, it harbors inherent limitations: tool invocation is essentially a stateless, memoryless atomic operation, incapable of sustaining deep comprehension of application business semantics and operational context; and the capability boundaries of tools are strictly delimited by predefined APIs, lacking flexibility and adaptive evolution capacity.

## 1.3 Main Contributions

This paper proposes a new integration paradigm, namely "Application-as-Agent". We construct a dedicated AI agent for each existing enterprise application to serve as its digital proxy, endowing it with comprehensive understanding of the application's internal structure, business logic, and data model, as well as autonomous navigation and intelligent manipulation capabilities.

Cross-application collaboration is subsequently achieved through natural language dialogue among agents, i.e., Conversation-as-Integration.

The central to our approach is the concept of Agentic Nesting, which we formally define as a hierarchical multi-agent organizational pattern characterized by three structural properties: (1) recursive encapsulation, where each enterprise application is wrapped as an autonomous agent with bounded capabilities; (2) layered composition, where individual application agents are nested within higher-order coordinating structures; and (3) emergent orchestration, where system-level behaviors arise from natural language negotiations among nested agents rather than predefined control flows. This nesting architecture distinguishes our framework from flat multi-agent topologies by introducing explicit hierarchical boundaries that mirror the organizational structure of enterprise application portfolios.

The main contributions presented herein are organized across three progressive levels, as shown in Table 1.

Table 1. Main Contributions

| **Level** | **Core Content** |
|---|---|
| **Extraction** | A standardized methodology for application agent initialization, capable of automatically extracting a fully functional AI agent from any existing enterprise application, thereby endowing systems originally lacking intelligent interaction capabilities with natural language dialogue and autonomous manipulation. |
| **Composition** | A multi-agent collaboration framework that organizes multiple application agents into a hierarchical collaborative network, achieving efficient coordination through central scheduling, knowledge sharing, and task negotiation mechanisms. |
| **Interaction** | A unified interaction interface centered on natural language, enabling users to accomplish cross-application data querying, process orchestration, and report generation through a single entry point, thereby realizing an interaction experience of "one conversation, global control". |

This paper further explores the generalization potential of this methodology in broader scenarios encompassing heterogeneous system coordination, and large-scale data management.

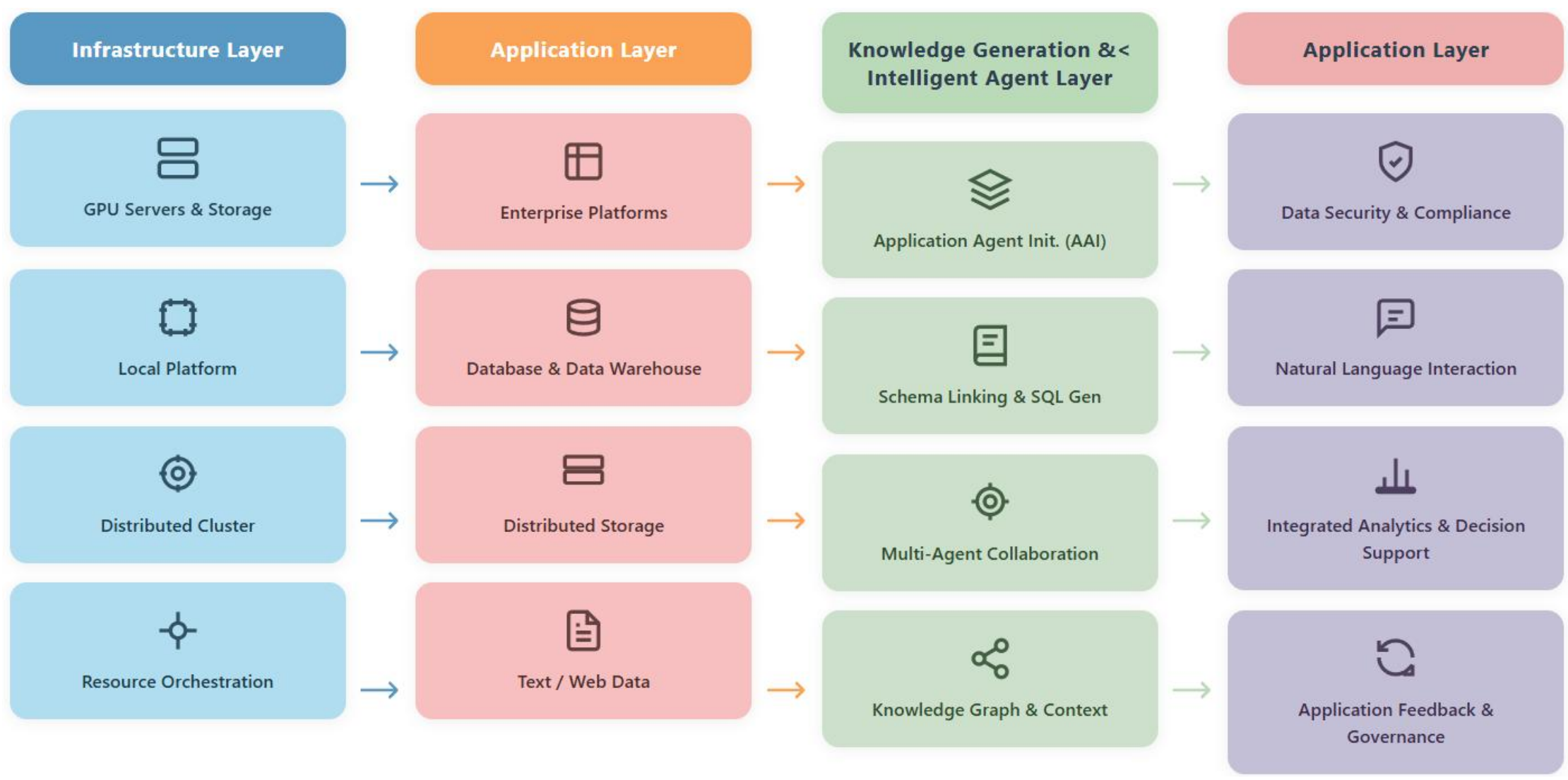

Figure 1. Enterprise Application-as-Agent Framework.

## 2. Related Work

## 2.1 Enterprise Application Integration

Enterprise Application Integration (EAI) as a distinct research discipline emerged during the early 1990s. Early approaches adopted point-to-point integration, wherein systems communicated directly through custom interfaces. However, as the number of systems grew, the number of interfaces increased at O(n2), causing maintenance costs to escalate dramatically. The Enterprise Service Bus (ESB) [22] introduced a middleware layer that reduced coupling through message routing and protocol conversion, yet the ESB itself became an architectural bottleneck with high configuration complexity.

The rise of microservices architecture has propelled the development of API gateway and service orchestration solutions [23]. Each microservice exposes standardized REST/gRPC interfaces, managed and routed through an API gateway. Nevertheless, this approach requires all systems to provide canonical APIs, which is a precondition that frequently fails for numerous legacy systems and third-party applications.

Integration Platform as a Service (iPaaS), exemplified by MuleSoft, Zapier, and Workato, offers low-code/no-code integration capabilities, lowering the barrier to entry. However, iPaaS still relies on predefined connectors and data mappings, necessitating substantial custom development when confronting non-standard systems.

Robotic Process Automation (RPA), exemplified by commercial platforms such as UiPath and Blue Prism, facilitates inter-system automation by emulating end-user interactions with graphical interfaces [24]. RPA does not depend on APIs and exhibits low invasiveness toward systems, yet its rule-based script-driven operations lack flexibility, rendering it fragile and brittle when interfaces change. Systematic analytical framework research has further revealed the structural challenges of RPA in terms of maintainability and scalability.

Researchers have further attempted to incorporate enterprise core data infrastructure directly into the agent collaborative network, transcending the dependency of traditional integration solutions on predefined interfaces. Wang et al. [25] proposed a multi-dimensional data analysis framework based on the interaction between LLM agents and Knowledge Graphs (KG), treating enterprise databases as the operational subjects of agents. Leveraging LLM agents, the framework automatically extracts structured knowledge from heterogeneous data sources and constructs domain knowledge graphs in real time, subsequently supporting users in deep exploration and intelligent analysis of data entities through an interactive visualization platform. This study reveals the dual value of knowledge graphs in enterprise data applications: on one hand, providing structured factual anchors for LLMs to mitigate hallucination risks; on the other hand, serving as the contextual environment and "workbench" for agents, supporting multi-dimensional analysis from macro-level correlations to micro-level insights. This "database-as-agent" perspective indicates that enterprise data assets can likewise transcend the passive response mode, being endowed with autonomous interaction capabilities and actively participating in collaborative analysis.

However, the aforementioned work primarily focuses on bidirectional agent–knowledge graph interaction in data analysis scenarios, and has not yet treated databases as general enterprise applications for systematic abstraction, nor explored their collaborative mechanisms within heterogeneous agent networks. The Application-as-Agent framework proposed in this paper does not require target applications to provide APIs or specific interfaces; instead, it relies on the semantic understanding and reasoning-planning capabilities of AI agents to achieve deep cognition, graph generation, and autonomous manipulation of applications. This paradigm shift fundamentally transcends the dependency of traditional integration solutions on predefined interfaces, offering a novel theoretical pathway for intelligent collaboration among heterogeneous systems.

## 2.2 AI Agent Frameworks

### 2.2.1 Single-Agent Frameworks

ReAct [8] proposes the paradigm of interleaving reasoning and acting, enabling large language models to dynamically invoke external tools during chain-of-thought reasoning processes, thereby realizing autonomous decision-making that thinks and acts concurrently.

Toolformer [9] further explores model autonomy in tool usage, employing self-supervised learning to enable models to learn when and how to invoke external tools to compensate for their own knowledge limitations. The Function Calling mechanism [26] has become the standardized interface specification for mainstream model providers including OpenAI, Anthropic, and Google, allowing agents to invoke external functions in a structured manner and significantly enhancing the systematicity and reliability of tool integration.

### 2.2.2 Multi-Agent Frameworks

MetaGPT [10] decomposes the full software engineering lifecycle into multiple specialized roles (e.g., product manager, system architect, development engineer), with each role enacted by an independent agent; cross-role collaboration and knowledge transfer are achieved through standardized documents. AutoGen [11] provides a flexible multi-agent conversational programming framework, supporting human-in-the-loop and asynchronous message-passing mechanisms among agents. ChatDev [27] simulates the organizational form of a virtual software enterprise, with multiple agents driving the complete software development workflow through natural language dialogue.

However, the core paradigm of the aforementioned multi-agent frameworks remains focused on task decomposition and role playing, the roles and capability boundaries of agents are predefined during the system design phase, and their knowledge sources are limited to the parametric knowledge of the large language model itself or pre-configured sets of external tools [12].

## 2.3 Agent Communication Protocols

The Model Context Protocol (MCP) [16], proposed by Anthropic, aims to provide standardized context management and tool access interfaces for large language models. MCP uniformly models external resources as "tools", with models accessing these resources through structured function call mechanisms. The protocol ecosystem has developed rapidly, generating numerous community-driven MCP Server implementations covering diverse application scenarios including databases, file systems, and web services.

The Agent-to-Agent (A2A) protocol [17], released by Google, endeavors to establish interoperability communication standards for agents built upon heterogeneous frameworks. A2A defines core abstractions including Agent Card (agent capability declaration), task lifecycle management, and asynchronous message passing, attempting to address the problems of discovery and collaboration among agents.

The Agent Communication Protocol (ACP) further explores more general agent communication specifications, seeking to establish unified interaction semantics within broader distributed agent networks.

Although these protocols have advanced the standardization process of the agent ecosystem, they still face several challenges for enterprise application development: (1) the lack of mature large-scale application scenarios; (2) agent capability descriptions relying on manual authoring, with information being insufficiently standardized and complete; (3) immature mechanisms for sharing and synchronizing memory and knowledge among multiple agents. Studies such as Reflexion demonstrate that agents can optimize memory and decision-making through self-reflection mechanisms [28]. Retrieval-augmented generation technology provides a scalable architecture for agents to dynamically acquire external knowledge [29].

Based upon critical assimilation of the design principles underlying the aforementioned protocols, this paper proposes a collaboration scheme oriented toward practical enterprise scenarios: rather than reducing external systems to atomic tools, it endows agents with deep cognitive capabilities toward systems through an application abstraction layer, thereby achieving autonomous collaboration grounded in semantic understanding rather than mechanical cooperation based on interface invocation.

# 3. Framework Design

The proposed Application-as-Agent framework is structured according to a hierarchical architectural model, organized in ascending order from the foundational Application Abstraction Layer, through the intermediary Agent Composition Layer, to the surface-level Interaction Layer.

The Application Abstraction Layer assumes the principal function of deriving and instantiating comprehensively operational application agents from pre-existing enterprise software systems. The Agent Composition Layer is responsible for constructing the multi-agent collaborative network, achieving task scheduling and knowledge sharing. The Interaction Layer is responsible for unified user access and result presentation. Figure 2 presents the conceptual interconnections and information trajectories characterizing the comprehensive system architecture.

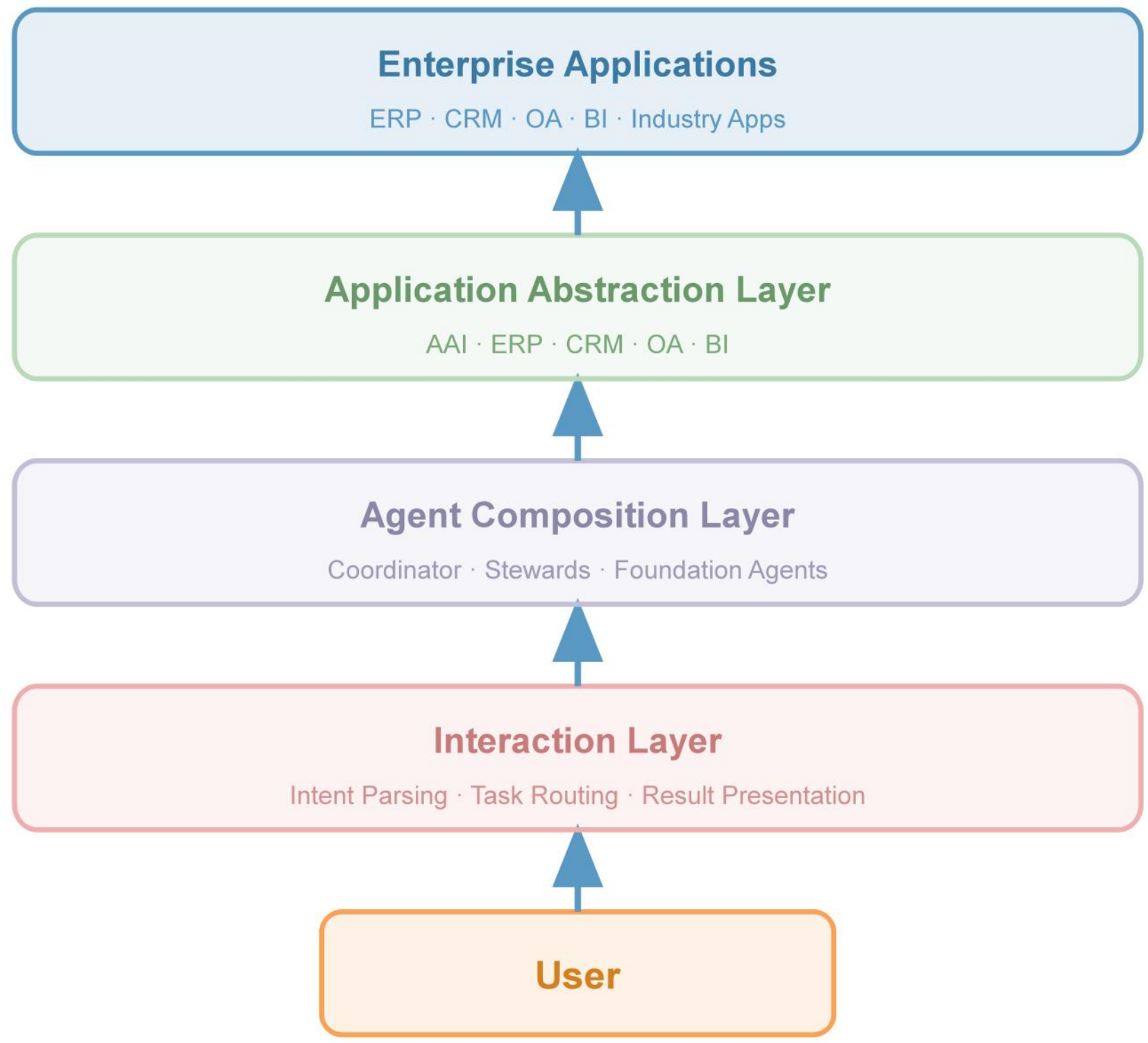

Figure 2. Layered Architecture of Application-as-Agent.

### 3.1 Application Abstraction Layer: From Application to Agent

The core task of the Application Abstraction Layer is to transform heterogeneous existing enterprise applications into AI agents, termed Application Accessories, possessing autonomous interaction capabilities. This process is defined as Application Agent Initialization (AAI), whose essence is an automated mapping mechanism from application ontology to agent ontology [30].

#### 3.1.1 Initialization Workflow

AAI follows a standardized workflow encompassing five stages: information collection, identity construction, skill generation, credential configuration, and capability verification.

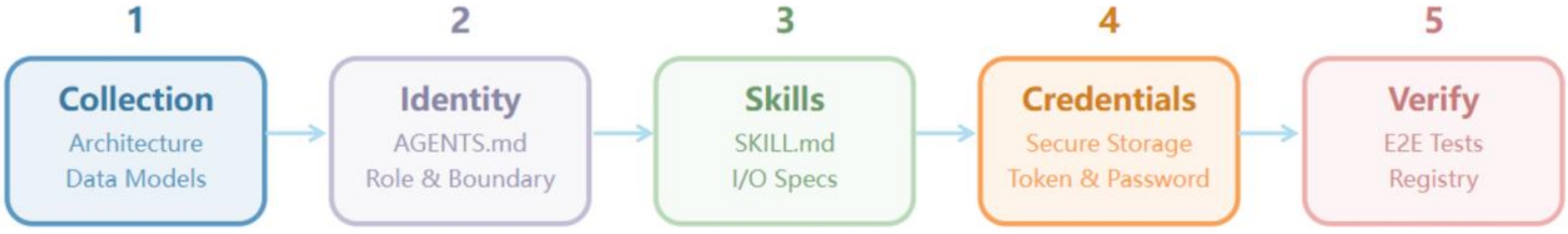

Figure 3. Application Agent Initialization (AAI) Workflow.

**Stage One: Application Information Collection.** The system automatically scans the technical architecture information of the target application; the information acquisition scope encompasses, without restriction to, the following elements: hierarchical directory organizations, underlying code architectures, exposed API service endpoints, underlying data schemas, user interface configurations, and modular functional components. For applications with backend code access privileges, the system further parses routing definitions, database models, and business logic layer implementations to construct a complete application semantic graph. On this basis, the system further extracts and constructs an Application Knowledge Graph from the collected multi-source information, explicitly depicting the application's functional modules, data entities, API interfaces, and their semantic associations in graph-structured form [31]. This knowledge graph not only provides agents with a structured panoramic description of the application but can also be dynamically injected as contextual input during subsequent human-agent dialogue, enabling agents to reason and respond based on precise understanding of the application topology, thereby significantly enhancing the accuracy and efficiency of interaction.

**Stage Two: Agent Identity Construction.** Based on the collected information, the system automatically generates an agent identity description file (AGENTS.md), specifying the following core attributes:

Role Definition: The application identifier represented by this agent and its domain of responsibility.

Capability Boundary: The set of executable operations and the list of high-risk operations requiring human confirmation.

Knowledge Scope: The semantic coverage of application modules, functions, and data resources mastered by the agent.

Interaction Protocol: The instruction reception format and result return specification.

**Stage Three: Skill File Generation.** The system creates structured skill description files (SKILL.md) for the agent; each skill entry explicitly defines: input parameter types and output format specifications, execution preconditions and permission requirements, invocation methods, and typical examples.

**Stage Four: Identity Credential Configuration.** Application access credentials (e.g., account passwords, API tokens) are configured in a secure manner, enabling the agent to access the target application with authorized identity. Credentials employ an independent encrypted storage mechanism, supporting secure handling of passwords containing special characters.

**Stage Five: Capability Verification and Registration.** End-to-end functional verification (e.g., authentication tests, data query tests) is performed on the initialized agent; upon confirmation of capability readiness, its capability declaration is registered with the central coordinating agent.

### 3.1.2 Initialization Constraints

The AAI process must satisfy the following critical constraints:

(1) Minimal Authorization Principle: The access privileges assigned to each agent should be rigorously constrained to the minimal operational scope necessitated by its designated functions. High-risk operations (e.g., data deletion, configuration modification) must incorporate secondary confirmation mechanisms.

(2) Application Compatibility: The framework must accommodate two scenarios: applications with backend access privileges and those with frontend access only. For applications with backend access, the agent can directly manipulate databases and APIs; for frontend-only applications, the agent achieves information acquisition through interface navigation and data extraction.

(3) Security Isolation: Each application agent operates within an independent working directory; its file system, environment variables, and execution context are mutually isolated to prevent cross-application data leakage.

(4) Idempotency Guarantee: Agent operations should be designed as idempotent whenever possible, enabling safe retry upon failure.

### 3.1.3 Capability Gains Post-Initialization

Upon completion of AAI, the application agent possesses the following hierarchical capability model:

(1) Read: Capable of browsing application interfaces, querying data, and extracting information.

(2) Review: Capable of inspecting and evaluating data and processes within the application.

(3) Edit: Capable of modifying configurations, data, and content within the application.

(4) Manage: Capable of executing application management operations, such as user management and permission configuration.

Critically, the AAI process endows legacy applications originally devoid of any AI interaction capabilities with natural language understanding and response capacities, this transformation itself constitutes a significant paradigm innovation. This transformation benefits from the development of pre-trained language models in deep contextualized semantic representation, as well as breakthroughs in the generative pre-training paradigm regarding in-context learning, reasoning, and code generation [32][33]. Through instruction fine-tuning techniques [34], models can accurately follow user intentions to execute complex tasks; via in-context learning mechanisms [35], agents can adapt to operation semantics in specific application domains without parameter updates; and prompt engineering methods [36] provide a systematic pathway for eliciting the latent capabilities of models. The maturation of open-source large language models (e.g., Llama 2 [37]) further lowers the technical threshold for constructing dedicated agents for heterogeneous applications. Users can directly engage in conversational interaction with the application agent in natural language to obtain information or issue instructions, independent of any prerequisite familiarity with the target application's particular operational procedures or underlying technical specifications.

## 3.2 Agent Composition Layer: From Individual to Collective

Upon completion of initialization for multiple application agents, it is necessary to further organize them into an agent network possessing collaborative decision-making capabilities. This process is defined as Agent Composition, whose core lies in establishing a hierarchical collaborative topology and dynamic task allocation mechanism.

### 3.2.1 Topology Structure

Definition 3.1 (Agentic Nest). An Agentic Nest is a hierarchical multi-agent organization denoted by the tuple $N = (A, L, \prec, \delta)$, where:

$A = \{a_1, a_2, ..., a_n\}$ is the set of application agents, each encapsulating a legacy enterprise application;

$L = \{l_coord, l_proxy, l_found\}$ is the set of stewardship layers, comprising central coordination, application proxy, and foundational capability;

$\prec \subseteq L \times L$ is a strict partial order representing delegation authority, with $l_proxy \prec l_coord$ and $l_found \prec l_coord$;

$\delta : T \times L \rightarrow 2^A$ is the dynamic task dispatch function mapping a task $\tau$ and layer l to a subset of agents capable of executing $\tau$ at layer l.

Property 3.1 (Nesting Completeness). An Agentic Nest N satisfies nesting completeness if, for every task $\tau \in T$, there exists a finite decomposition $\tau = \cup_{i=1}^{k} \tau_i$ such that each subtask $\tau_i$ is executable by some agent $a \in A$ at its assigned layer $l(a) \in L$.

This nested architecture embodies the core principle of Agentic Nesting: lower-level application agents remain encapsulated within their respective domains, while higher-level coordinating agents possess global visibility but delegate execution authority downward. Such nested delegation ensures that complexity remains localized, and system-wide coordination emerges through recursive composition rather than centralized control. The framework adopts a hierarchical topology of "central scheduling + application steward":

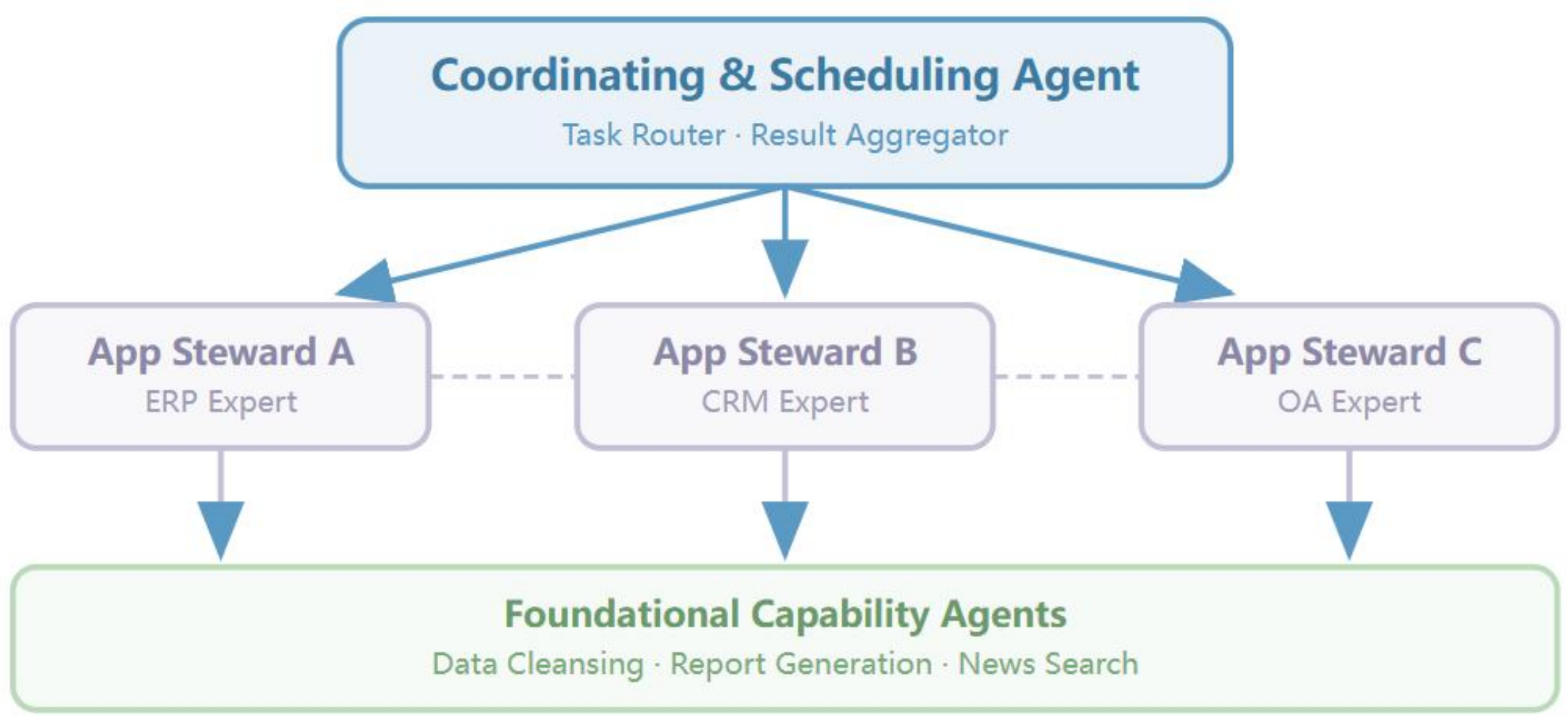

Figure 4. Hierarchical Structure.

As shown in Figure 4, the proposed topological structure encompasses three distinct functional responsibilities:

**Coordinating and Scheduling Agent:** Serving as the global task router and result aggregator. Upon receiving user instructions, it parses intentions, decomposes them into subtasks, dispatches them to appropriate sub-agents, and finally aggregates results for return to the user. The coordinating agent does not directly access any application; its exclusive function pertains to orchestration and collaborative mediation.

**Application Steward Agent:** Each application corresponds to one steward agent, serving as the deep expert of that application. The steward agent comprehends the application's full functionality, data structures, and operational workflows, capable of independently completing complex tasks within the application.

**Foundational Capability Agent:** Providing general services (e.g., data cleansing, report generation, news search) that can be invoked by any application steward or coordinating agent.

The advantages of this topology structure are as follows: it decouples task scheduling from task execution, thereby permitting the coordinating agent to concentrate on abstract semantic comprehension and strategic orchestration, while enabling the application proxy agent to specialize in domain-specific knowledge utilization and concrete operational implementation, and the foundational capability agent to focus on general capability reuse, thereby forming clear responsibility boundaries and an extensible collaboration paradigm [38].

### 3.2.2 Agent Design Principles

The multi-agent collaborative network conforms to four foundational design tenets aimed at guaranteeing operational dependability, semantic coherence, and collaborative security.

(1) Persona Independence requires each agent to possess an independent identity identifier, role positioning, and behavioral pattern, strictly distinguishing its own responsibilities from the functional boundaries of other agents, preventing role confusion and responsibility drift, thereby ensuring the predictability of agent behavior and clarity of responsibilities.

(2) Knowledge Sharing stipulates that public knowledge, including industry domain knowledge, terminology dictionaries, and analysis rules, be uniformly stored and shared across the agent network, avoiding redundant learning and knowledge redundancy among individual agents and enhancing the overall knowledge utilization efficiency and semantic consistency of the network.

(3) Memory Isolation ensures that each agent independently maintains its session history and contextual memory states; task execution by other agents will not interfere with or contaminate this agent's memory space, safeguarding contextual integrity and reasoning continuity in multi-task concurrent scenarios.

(4) Authority Autonomy grants each application proxy agent exclusive manipulation authority over the application resources under its responsibility; all other agents are strictly forbidden from accessing or manipulating these application resources without explicit authorization, thereby implementing both the minimal authorization principle and rigorous application-level security compartmentalization.

These four principles collectively constitute the governance framework for multi-agent collaboration, achieving orderly group collaboration while safeguarding individual autonomy, providing institutional guarantees for the stable operation of large-scale agent networks [39].

### 3.2.3 Communication Mechanism

Inter-agent communication employs natural language dialogue as the core interaction medium, rather than traditional structured API calls. This design choice is grounded in three theoretical considerations.

Regarding expressive flexibility, natural language can convey subtle semantic information including ambiguous requirements, priority implications, and context dependencies. Its expressive capacity far exceeds the formal constraints of structured parameter sets, rendering task descriptions more aligned with human cognitive habits.

Regarding evolutionary robustness, when the capability boundaries of target agents undergo dynamic changes, natural language invocation merely requires adjustment of description strategies, without necessitating modification of interface definitions or redeployment of contracts, significantly reducing coupling costs and adaptation overhead during system evolution.

Regarding audit explainability, natural language dialogue records among agents inherently constitute audit trails for task execution, facilitating post-hoc tracing, responsibility attribution, and process review, satisfying enterprise-level compliance and governance requirements.

At the implementation level, the scheduling agent can establish communication with lower-layer agents through three modes. The first is CLI Agent Mode, wherein commands are executed and outputs captured within the target agent's independent working directory through a CLI Agent runtime (e.g., OpenCode [40], Claude Code), applicable to scenarios requiring direct file system manipulation or local tool execution. The second is SDK API Mode, wherein standardized programming interfaces are employed to create session instances and send structured messages, applicable to scenarios requiring deep integration with existing application systems. The third is Protocol Interop Mode, wherein standardized agent communication protocols such as the Agent Communication Protocol (ACP) are utilized to achieve cross-framework interoperability, applicable to cross-application collaboration scenarios among heterogeneous agent frameworks. These three modes can be flexibly configured according to deployment environments and security policies, forming a stratified and progressive communication capability spectrum that provides adapted communication infrastructure for enterprise application scenarios of varying complexity.

## 3.3 Interaction Layer: From Collective to User

The Interaction Layer serves as the unified interface between users and the multi-agent collaborative network; its core function lies in shielding users from the complexity of underlying agent collaboration while providing a natural, coherent conversational interaction experience. Figure 5 shows the implementation of the interaction interface, where users access multiple heterogeneous enterprise applications through a single conversational entry point.

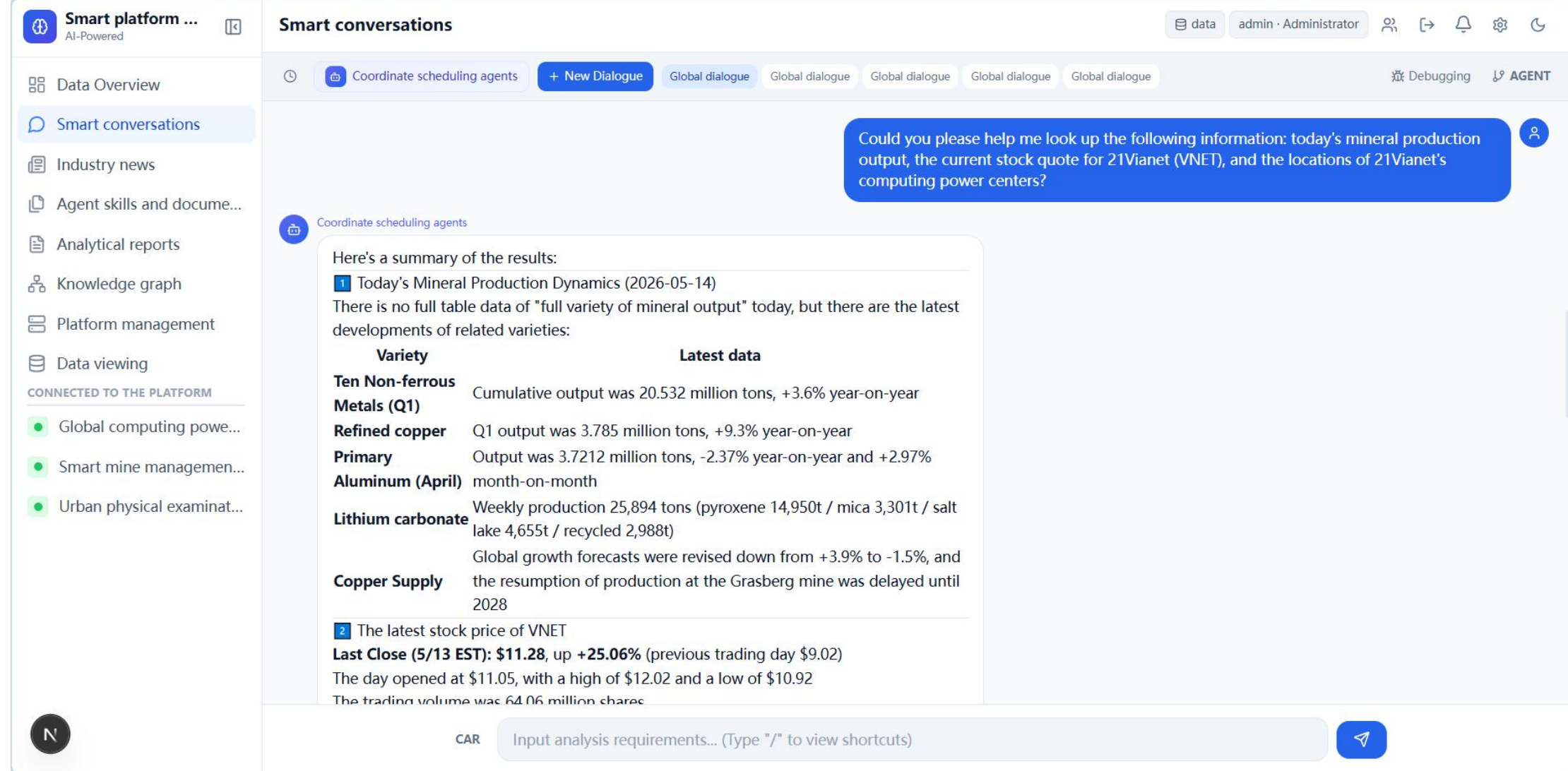

Figure 5. The Conversational Interaction Interface of Enterprise Application Multi-agent Platform.

### 3.3.1 Intent Parsing and Task Routing

Natural language instructions input by users are first subjected to intent parsing by the coordinating agent. Combining current session context with historical interaction records, the coordinating agent determines the target applications involved in the task and the required capability scope, subsequently generating a dynamic execution plan. Studies such as LLM+P demonstrate that transforming natural language goals into formal planning problems can endow large language models with optimal planning capabilities [41], providing methodological support for the coordinating agent's task decomposition and path optimization. Based on task complexity and cross-application dependency relationships, task routing follows three strategies.

First, for single-application tasks, the coordinating agent directly routes instructions to the corresponding application proxy agent, which independently completes execution as a domain expert and returns results.

Second, for multi-application tasks, the coordinating agent decomposes the task into subtasks with explicit input-output contracts based on task semantics, dispatches them to different application proxy agents for parallel execution. HuggingGPT demonstrates the potential of coordinating multimodal tools through language models to solve complex AI tasks [42], whose task planning and model selection mechanisms provide a reference paradigm for the coordinating agent's cross-agent scheduling. Upon completion of all subtasks, the coordinating agent performs result aggregation and consistency verification.

Third, for general-purpose tasks, the coordinating agent routes them to foundational capability agents (e.g., report generation, data analysis), leveraging cross-application general services to complete processing.

### 3.3.2 Result Aggregation and Presentation

After collecting returned results from sub-agents, the coordinating agent performs multi-level result aggregation and presentation optimization.

At the data level, the coordinating agent executes merging, deduplication, and format unification operations on multi-source returned results, eliminating semantic conflicts and structural discrepancies among heterogeneous data.

At the expression level, the coordinating agent generates structured reports, visualized charts, or natural language summaries based on user intent and scenario requirements, achieving multimodal information presentation. Multi-agent debate mechanism research demonstrates that discussion and consensus formation among independent agents can significantly enhance the factuality and reasoning reliability of results [43], providing theoretical reference for the coordinating agent's consistency verification.

At the provenance level, the coordinating agent annotates each data item in the aggregated results with its source application identifier and confidence score, ensuring result traceability and verifiability, satisfying the stringent requirements for information reliability in enterprise-level decision-making scenarios. Meanwhile, training data extraction attack research indicates that agent systems must establish strict input validation and privacy protection mechanisms [44], to prevent potential leakage risks of sensitive information during cross-application transit.

# 4. Implementation

This section presents a multi-application intelligent analysis system as a concrete case study to elaborate the detailed engineering implementation and technical specifics of the Application-as-Agent framework.

## 4.1 System Architecture

The system adopts an overall layered technical architecture. The frontend is built upon Next.js (React) for web application construction, furnishing three distinct functional perspectives: a dialogue-based interaction interface, an administrative management console, and a data visualization control panel.

The backend API is implemented via Next.js API Routes, assuming core responsibilities including identity authentication, application lifecycle management, agent scheduling, and data persistence.

The Agent executor adopts a pluggable architecture design, supporting multiple underlying runtime implementations and achieving runtime independence through a unified interface abstraction.

The data storage layer employs JSON and CSV files as lightweight persistence solutions, while reserving database backend interfaces to support large-scale deployment scenarios.

Crucially, the physical deployment topology materializes the Agentic Nesting principle defined in Section 3.2.1: the coordinating agent executes within the orchestration tier, maintaining global visibility over the agent registry but possessing no direct access to any application backend; application proxy agents are deployed within isolated working directories at the stewardship tier, each encapsulating a single legacy application; foundational capability agents operate as shared services at the base tier, accessible to both the coordinating agent and application proxy agents. This three-tier physical nesting mirrors the logical delegation hierarchy $\prec$ defined in Definition 3.1, ensuring that complexity remains localized within each agent's bounded context and that cross-layer communication is strictly mediated by the scheduling middleware.

## 4.2 Agent Runtime Adaptation

The framework adheres to a tool-agnostic design principle, shielding underlying runtime differences through a protocol adaptation layer so that upper-layer scheduling logic need only concern itself with target agent identifiers and message content, without perceiving specific runtime implementations.

Within the CLI Agent runtime category, the system supports three mainstream implementations. The first is OpenCode [40], a Go-based open-source AI coding assistant that supports launching NDJSON streaming output via the run format JSON parameter, applicable to automated scheduling and batch processing scenarios. The second is Claude Code [45], the official CLI coding tool released by Anthropic, which supports obtaining structured JSON output through the print-output format JSON parameter. The third is the Cursor Agent SDK [46], a programmatic agent interface provided by the Cursor IDE, supporting deep integration and custom extension.

Within the API-Based Agent category, the system supports custom agents built upon the LangChain or LangGraph frameworks, exposing capability interfaces through HTTP APIs. It is also compatible with native interface protocols such as the OpenAI Assistants API and the Anthropic Messages API, achieving seamless integration with mainstream large model service providers.

The protocol adaptation layer encapsulates the invocation differences of the aforementioned heterogeneous runtimes through a unified abstract interface, providing consistent semantics for agent creation, message sending, and result reception to upper layers, significantly reducing the coupling and maintenance costs of scheduling logic.

## 4.3 Application Initialization Implementation

The technical implementation of the application initialization module encompasses two core stages: directory structure generation and automatic document generation.

In the directory structure generation stage, the system automatically generates standardized working directory structures based on application type (with or without backend access privileges). This structure contains necessary components including agent identity files, skill description files, secure credential storage areas, and log audit directories, guaranteeing both the uniformity and strict segregation of execution environments allocated to individual application agents.

In the automatic analysis and document generation stage, the system employs AI agents to automatically scan the code repository or interface structure of target applications, generating comprehensive reference documentation. This documentation covers four dimensions: (1) technical architecture description, elaborating the application's overall technology selection and component dependency relationships; (2) functional module inventory, enumerating the core business functions and operation entry points provided by the application; (3) API endpoint list, if backend access privileges are available, comprehensively cataloging invocable service interfaces; (4) data model description, parsing the application's data entity relationships and field semantics. The aforementioned documentation constitutes the core knowledge base for application agents, providing the semantic foundation for their subsequent task comprehension and autonomous operation.

## 4.4 Agent Scheduling Implementation

The scheduling of subordinate applications is facilitated via a consolidated scheduling middleware, functioning as an intermediate mediation stratum positioned between the central coordinating agent and the respective application proxy agents, responsible for task dispatch and result return. The middleware operationalizes the nested delegation model central to Agentic Nesting: rather than permitting the coordinating agent to directly invoke application internals, the middleware enforces a strict downward-routing policy whereby all dispatches from the coordination layer must traverse this intermediary before reaching application proxy or foundational capability agents. This guarantees encapsulation boundaries that coordinating agents orchestrate via intent parsing and result aggregation alone, while application proxy agents retain exclusive manipulation authority over their respective application resources. Consequently, the scheduling layer serves as the runtime enforcement mechanism for the authority autonomy and memory isolation principles articulated in Section 3.2.2.

The scheduling workflow follows five stages.

**Stage One: Configuration Parsing.** The scheduling script retrieves application configuration information from the agent registry based on the target application name, including metadata such as working directory paths, agent runtime types, and version requirements.

**Stage Two: Environment Initialization.** The scheduling script initializes an agent runtime process within the application-dedicated working directory, establishing an independent execution context.

**Stage Three: Instruction Delivery.** The scheduling middleware transmits end-user directives or decomposed subtask specifications to the designated target agent utilizing formally structured message formats.

**Stage Four: Streaming Response.** The scheduling script receives the agent's execution process data and intermediate results in real time through standard output streams.

**Stage Five: Result Return.** The scheduling script returns aggregated execution results to the caller, which may be an upper-layer coordinating agent or the interaction layer directly facing end users. This design aligns with the agent-centric scheduling paradigm proposed in recent heterogeneous agentic systems research, where an intermediary scheduling layer decouples task admission from execution to minimize end-to-end latency across diverse runtimes [48].

## 4.5 Streaming Communication Protocol

The system adopts NDJSON (Newline Delimited JSON) as the streaming communication protocol for agent execution processes. This protocol stipulates that each line of data constitutes an independent JSON object, carrying standardized event type identifiers and data payload fields, facilitating frontend parsing and incremental rendering.

The frontend application achieves real-time visualized presentation of agent execution progress through NDJSON stream subscription. The displayed content encompasses four dimensions: currently executing tool invocation information, including tool names, input parameters, and execution status; incremental text output, presenting the agent's reasoning process and intermediate conclusions character-by-character via typewriter effect; token consumption and cost statistics, providing real-time feedback on computational resource usage and estimated costs; and task completion status indicators, explicitly distinguishing three terminal states, including execution success, partial success, and failure, along with accompanying error diagnostic information.

## 5. Case Study

To empirically assess the efficacy of the proposed Application-as-Agent paradigm within intricate multi-system operational contexts, this paper deploys and evaluates the framework in cross-industry multi-application analysis scenarios. The empirical evaluation employs two distinct enterprise application platforms characterized by divergent technical architectures: a Global Computing Power and New Energy Electricity Information Platform (subsequently denoted as the "Energy-DC Platform") alongside an Intelligent Mining Operations Management Platform (designated Mining 2.0). The former is a GIS-based global data center and new energy electricity information visualization and analysis platform, covering computing center spatial distribution, new energy installed capacity data, power structure trends, and industry reports. The latter is an AI+GIS+RS-based intelligent mining decision management platform, encompassing mining truck GPS real-time tracking, trinity energy management (photovoltaic, energy storage, charging piles), safety monitoring and alerting, and remote sensing interpretation. Figure 6 presents the typical interfaces of the two experimental platforms, intuitively illustrating the heterogeneity of the target systems.

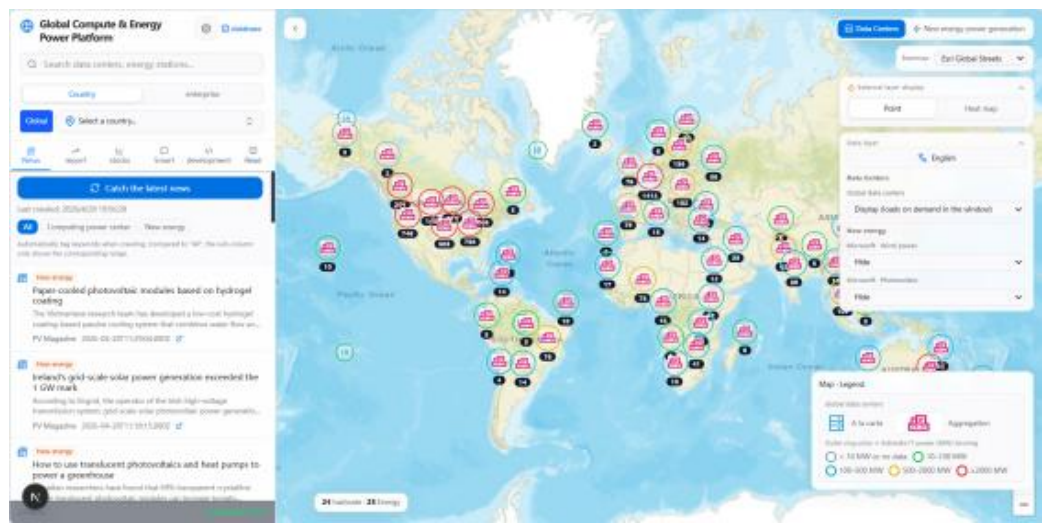

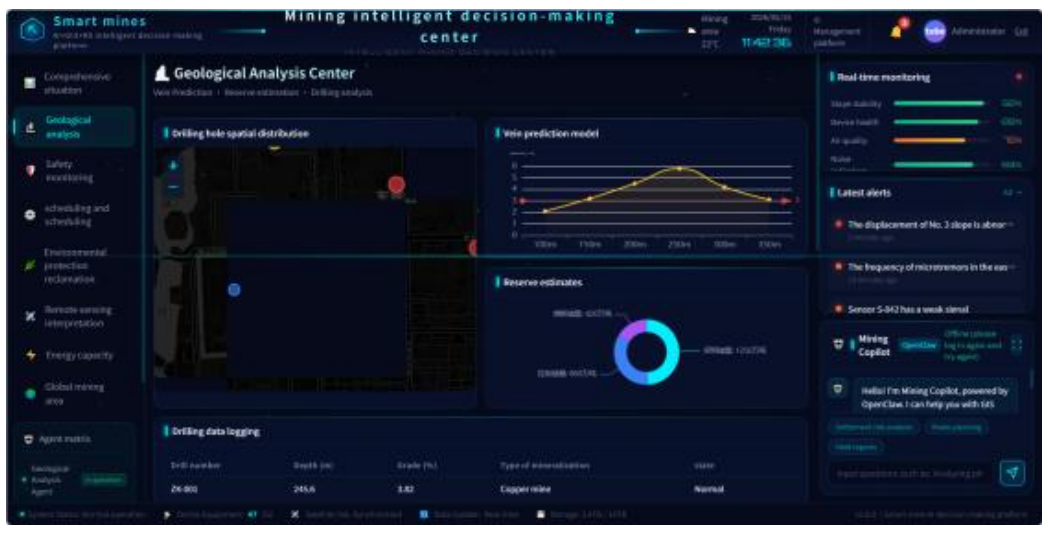

Figure 6. Experimental Platform Interfaces. (Left) The GIS visualization interface of the Energy-DC Platform, displaying the spatial distribution of computing centers in Central Asia and the new energy electricity heatmap. (Right) The data dashboard of the Mining 2.0 Platform, showing real-time mining truck locations, trinity energy management, and safety monitoring situational awareness.

The two platforms employ heterogeneous technical architectures (Next.js/React + SQLite/Spatialite versus AI + GIS + RS + Leaflet/ECharts), data models (global computing center spatial data versus mining real-time IoT operational data), and communication protocols (REST API/WebSocket versus H5 GPS reporting/WebSocket), covering three typical task categories: structured spatial data querying, real-time IoT data retrieval, and cross-system process orchestration.

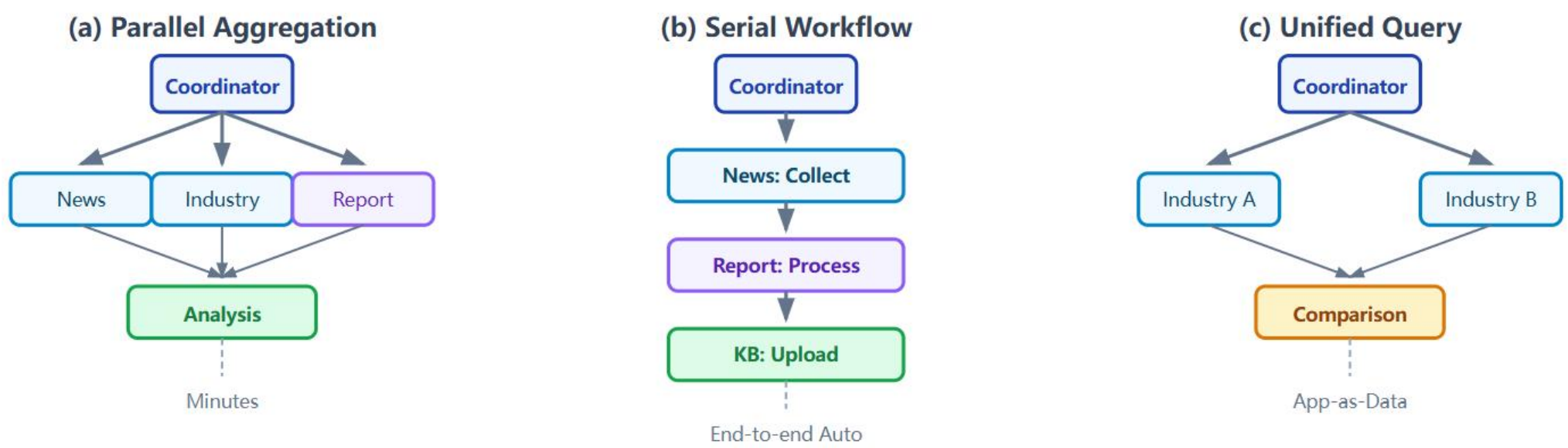

Figure 7. Case Study Scenarios.

## 5.1 Scenario One: Multi-Source Heterogeneous Data Aggregation and Intelligent Analysis

This scenario validates the framework's capabilities in multi-source information aggregation and automated report generation. The user inputs an instruction via natural language dialogue, requesting a comparative analysis of new energy supply structures for computing centers in Central Asia and energy management efficiency in smart mining operations, with generation of a structured analysis report.

Upon receiving this high-level instruction, the coordinating agent performs intent parsing and task decomposition, breaking it down into four subtasks with explicit data dependency relationships. The first subtask is executed by the news agent (a foundational capability agent), responsible for retrieving recent news and policy updates regarding new energy development, data center expansion, and green mining transformation in Central Asia, extracting key event information through keyword matching and semantic filtering. The second subtask is executed by the Energy-DC platform proxy agent, which logs into the platform with authorized credentials to obtain multi-dimensional spatial and attribute data, including the distribution of computing centers in Central Asia, new energy type proportions (solar, wind, hydro), Power Usage Effectiveness (PUE) metrics, and installed capacity. The third subtask is executed by the Mining 2.0 platform proxy agent, which accesses operational data including real-time photovoltaic array power, energy storage State of Charge (SOC), charging pile utilization rates, and mining truck energy consumption and haulage volumes. The fourth subtask is executed by the report generation agent, responsible for aggregating the heterogeneous information returned by the news agent and the two application proxy agents to generate a structured analysis report containing energy structure comparison charts, efficiency trend analysis, and policy interpretation.

This scenario prominently demonstrates the efficiency advantages of the framework. Under traditional modes, analysts must manually switch among multiple independent systems, including GIS map interfaces, database management backends, mining data dashboards, and document editing tools, to perform data retrieval, format conversion, and information summarization, typically requiring several hours to complete [49]. Under the Application-as-Agent framework, the user need only issue a natural language instruction through a single conversational entry point; the entire aggregation and analysis workflow is automatically completed within minutes, and the output results possess consistency and traceability.

## 5.2 Scenario Two: Cross-Application Automated Workflow

This scenario validates the framework's capabilities in cross-system process chaining and automated execution. The user instruction requests completion of a serial task involving three heterogeneous systems: collecting current-day news and operational data from both the energy and mining domains, generating a standardized operations brief, and uploading it to the enterprise internal knowledge base application.

The coordinating agent recognizes the cross-system nature of this instruction and the serial dependency relationships among tasks, subsequently generating a three-phase execution plan. Phase one is executed by the news agent and the Energy-DC platform proxy agent in parallel: the news agent aggregates current-day industry news from external news sources, while the Energy-DC proxy agent extracts the latest industry reports and key indicators (e.g., electricity consumption trends, power structure distributions) from the platform. Simultaneously, the Mining 2.0 platform proxy agent retrieves mining operational logs including mining truck dispatch status, safety alert events, and energy equipment operation records. Phase two is executed by the report generation agent for content processing, transforming the raw news materials and multi-source operational data into a standardized operations brief format conforming to enterprise specifications. Phase three is executed by the internal application proxy agent for system operations, logging into the enterprise knowledge base application with authorized credentials to upload the generated brief file to the designated directory and complete metadata annotation.

This scenario reveals the structural inefficiency of traditional integration modes. In conventional workflows, the aforementioned tasks require operators to separately manipulate news search engines, the Energy-DC GIS visualization interface, the Mining 2.0 management backend, and document editing tools, with humans assuming the inter-system interfacing and coordination functions[49]. The Application-as-Agent framework replaces manual process chaining through autonomous collaboration among agents, realizing comprehensive automation across heterogeneous system boundaries while substantially mitigating operational intricacy and the propensity for human-induced inaccuracies.

### 5.3 Scenario Three: Application as Data—Unified Cross-Platform Data Querying

This scenario validates the framework's capabilities in unified access and comparative analysis of heterogeneous data across the two platforms. The user instruction requests comparison of energy efficiency fluctuation trends between computing centers and mining operations within a specific time period, a task involving energy consumption data stored in different applications with heterogeneous data models.

The coordinating agent recognizes the parallel nature of this task and simultaneously schedules the Energy-DC platform proxy agent and the Mining 2.0 platform proxy agent to perform data extraction. Acting as deep domain experts of their respective applications, the Energy-DC proxy agent extracts historical PUE metrics, new energy generation volumes, and electricity consumption trends for computing centers from the spatial database (datacenter_spatial.db), while the Mining 2.0 proxy agent extracts photovoltaic generation volumes, energy storage charge/discharge volumes, charging pile utilization, and mining truck energy consumption trends from the mining operational database. Both agents return structured results carrying application identifiers and data schema descriptions. Upon receiving multi-source heterogeneous data, the data processing agent performs data cleansing, format standardization, and semantic alignment operations, unifying energy consumption units, temporal granularities, and metric definitions, to eliminate data definition discrepancies and dimensional conflicts between the two platforms. Based on the standardized unified dataset, the report generation agent creates comparative analysis charts and textual interpretations, intuitively presenting the energy efficiency characteristics and correlations between the computing power and mining industries.

This scenario embodies the framework's core philosophy of "connecting applications means connecting data". Under the Application-as-Agent paradigm, each enterprise application is regarded as an autonomous data source node. Through the semantic abstraction and standardized interfaces of application agents, data resources dispersed across heterogeneous systems, including spatial computing center databases and real-time mining IoT data, are unified into the agent network. End-users are not required to possess knowledge regarding the precise physical data repositories, communication protocols, or underlying schema configurations. They need only express analysis intentions through natural language to achieve transparent cross-application data querying and comparative analysis.

## 6. Discussion

### 6.1 The Significance of Endowing Applications with AI Capabilities

The transformative characteristic of the Application-as-Agent framework lies in its endowment of existing enterprise applications with autonomous intelligent interaction capabilities. This capability gain is not merely a simple superposition of technical functionalities; rather, it denotes a substantive transformation in human-computer interaction modalities, transitioning from Graphical User Interface (GUI)-centric paradigms toward conversational natural language interfaces. Early experiments with GPT-4 reveal its emergent capabilities across multiple domains including mathematical reasoning, code generation, and visual understanding [50], providing the capability foundation for the maturation of the natural language interaction paradigm.

Under traditional operational modes, users are required to individually learn the specific interface layouts, interaction logic, and operational paths of each application, with cognitive burden increasing linearly with the number of applications. Under the Application-as-Agent paradigm, all heterogeneous applications establish interaction with users through a unified natural language interface, producing threefold effects. First, user learning costs are substantially reduced that users need only describe business requirements in natural language, without memorizing complex operational steps and interface navigation paths. Second, the usage threshold of applications is transformed from operational skill requirements to expressive capability requirements, extending access privileges to complex professional systems from technical personnel to the full spectrum of business personnel. Third, users without technical backgrounds can efficiently utilize enterprise systems that originally required professional training to operate, achieving the realization of Technology Democratization within the organization.

### 6.2 A New Approach for Application Integration

From the theoretical perspective of enterprise application integration, Application-as-Agent provides a fundamental paradigm alternative. Traditional EAI solutions employ interfaces as the basic unit of integration, with systems exchanging structured data through predefined API contracts, and integration depth is constrained by interface expressiveness and contract stability. Application-as-Agent, conversely, employs agents as the basic unit of integration, with systems achieving dynamic collaboration through semantic dialogue; integration depth is determined by the cognitive capabilities of agents rather than interface definitions.

This paradigm shift yields threefold advantages. In the dimension of data unification, application proxy agents can comprehend the data semantics and business contexts of different applications, achieving deep semantic fusion beyond format conversion, rather than merely remaining at the level of field mapping and protocol adaptation. In the dimension of process unification, complex cross-system business processes no longer require pre-orchestrated workflow engines to define execution paths. The coordinating agent can dynamically plan optimal execution strategies based on real-time context, possessing adaptive capabilities to respond to anomalies and changes. In the dimension of management unification, enterprise managers can obtain operational status overviews of multiple heterogeneous systems through natural language dialogue, achieving cross-system situational awareness and coordinated decision-making, transcending the traditional management cockpit's dependency on predefined metrics and fixed views.

### 6.3 The Application-as-Data Paradigm

We introduce the conceptual framework of "Application-as-Data", predicated on the fundamental premise that establishing connectivity with an application inherently provides access to the complete spectrum of data assets encapsulated within that application. When multiple heterogeneous applications achieve interconnection through the agent network, the system naturally forms a distributed Federated Data Query Network, enabling logical-level unified data access without centralized physical data migration.

This concept possesses three implications. Primarily, organizations are no longer obligated to allocate considerable capital expenditures toward developing centralized data repositories or extensive data lake infrastructures for the physical consolidation of information from disparate sources; application proxy agents can extract required information on demand from each data source in real time, achieving architectural optimization of computation migrating to data rather than data migrating to computation. The architectural progression observed in NewSQL database systems offers pertinent engineering insights regarding the reconciliation of horizontal scalability requirements with strong consistency guarantees within distributed computing environments [51]. Second, the data governance model shifts from the traditional "extract-cleanse-store" paradigm to the "in-situ query-intelligent aggregation-on-demand delivery" paradigm, reducing the complexity and latency costs of maintaining data consistency. Third, application management capabilities and data management capabilities form an isomorphic mapping that permission configuration, access control, and audit strategies for managing applications are directly transformed into data governance infrastructure, avoiding compliance risks caused by the disconnection between application management and data governance in traditional modes.

Furthermore, if the abstraction object of the Application-as-Agent framework is extended from enterprise applications to Database Management Systems (DBMS), it becomes possible to construct intelligent data applications possessing unified multi-database querying capabilities. In this extended architecture, each database instance is managed by a dedicated database proxy agent that masters the database's schema structure, query language, and optimization strategies. Users express cross-database querying intentions through natural language; the coordinating agent automatically decomposes the query plan and schedules each database proxy agent to execute distributed queries, ultimately aggregating heterogeneous results into a unified response. This extension validates the universality and generalization potential of the Application-as-Agent methodology in the domain of data infrastructure.

## 6.4 Comparison with the Tool-Use Paradigm

Contemporary mainstream agent frameworks generally adopt the Tool-Use paradigm, modeling external systems as invocable tool sets through standardized interfaces such as the Model Context Protocol (MCP). The Application-as-Agent framework exhibits essential differences from this paradigm, which can be systematically compared across five dimensions, as shown in Table 2.

Table 2. The comparison between Tool-Use Paradigm and Application-as-Agent framework.

| Dimension | Tool-Use Paradigm (MCP/Tool-Use) | Application-as-Agent (Conversation-as-Integration) |
|---|---|---|
| System Positioning | Applications passively serve as tools to be invoked | Applications actively participate as agents in collaboration |
| Capability Boundaries | Constrained by predefined APIs/functions | Agents can autonomously explore and exploit application capabilities |
| State Management | Stateless atomic invocations | Stateful, context-aware sustained interactions |
| Adaptability | Interface changes necessitate tool definition modifications | Agents can self-adapt to application changes |
| Complex Tasks | Require precise function call chains | Complex tasks can be accomplished through dialogue descriptions |

The two paradigms are not mutually exclusive; rather, they represent complementary approaches suited to different scenario characteristics. The Tool-Use paradigm is more appropriate for highly structured, deterministic simple operation scenarios, offering explicit execution paths, controllable latency, and predictable costs. The Application-as-Agent paradigm is more suitable for scenarios requiring deep semantic understanding, contextual reasoning, and complex multi-step coordination, possessing the capacity to accommodate imprecisely specified requirements and adapt to operationally dynamic contexts. In practical engineering deployment, the two can form a layered complementary architecture: Tool-Use is employed for standardized, high-frequency atomic operations to ensure efficiency, while agent dialogue is adopted for complex, variable business processes to ensure flexibility.

## 6.5 Limitations and Challenges

This paper candidly acknowledges the limitations of the current framework at both the engineering practice and theoretical refinement levels, categorizing them into the following four dimensions.

First, latency and cost constraints. Natural language dialogue invocations among agents incur higher communication latency and computational costs compared to direct API function calls, primarily attributable to the inference overhead of large language models and the accumulation of dialogue turns. In scenarios with stringent real-time requirements (e.g., high-frequency trading, real-time control), this constraint may become a bottleneck for system deployment, necessitating further optimization of communication granularity and caching mechanisms among agents.

Second, reliability assurance. The inherent stochasticity in large language model generation processes may engender discrepancies in the operational determinations executed by application proxy agents. Large language model evaluation research has established a multi-dimensional evaluation system encompassing knowledge, reasoning, safety, and bias, indicating that agent systems must establish more rigorous assurance mechanisms regarding factuality, consistency, and safety. For high-risk operations involving fund transfers, permission modifications, and similar activities, although the current framework has incorporated secondary confirmation mechanisms, more systematic human review workflows, operation rollback mechanisms, and error recovery strategies must be established to safeguard the security baseline of enterprise applications. Adversarial sample research reveals the vulnerability of neural networks to input perturbations, providing a risk framework for input validation and anomaly detection in agent systems.

Third, scalability bottlenecks. As the number of connected applications grows, the decision space for task scheduling by the coordinating agent expands exponentially; the accuracy of intent parsing and the optimization efficiency of task allocation may consequently degrade. Future research must explore more efficient multi-agent routing algorithms, distributed scheduling architectures, and hierarchical coordination mechanisms to support the stable operation of large-scale agent networks.

Fourth, security and compliance governance. Application proxy agents possess operational privileges over the applications they manage; the security and compliance of their behavior directly impacts enterprise data assets and business continuity. Although the current framework has achieved working directory isolation and least-privilege configuration, a more comprehensive agent behavior auditing system, anomaly detection mechanisms, and compliance verification frameworks must be established to ensure that agent operations remain fully traceable, auditable, and controllable throughout their lifecycle.

## 7. Future Work

Drawing upon the theoretical investigations and empirical engineering validations conducted within the present framework, this paper outlines the following five research directions to further refine the methodological system and system capabilities of Application-as-Agent.

**Multi-Entity Initialization.** The current application agent initialization workflow is primarily designed for software applications; future work should design differentiated agent extraction strategies for various types of underlying entities, including hardware devices, mobile applications, and database management systems. For instance, for IoT devices, proxy agents with device control capabilities can be automatically generated by parsing their control code and communication protocols; for mobile applications, application functional semantic graphs can be constructed through interface reverse analysis and API probing, subsequently generating application proxy agents. This direction aims to establish a universal initialization methodology covering heterogeneous entity types, expanding the applicability scope of Application-as-Agent.

**Multi-Agent Negotiation Protocol Optimization.** The current multi-agent negotiation mechanism adopts a simple task distribution and result aggregation model; future work need to address more efficient collaboration protocols. This encompasses three sub-directions: memory sharing and knowledge synchronization strategies among agents, exploring cross-agent context transfer and experience reuse while preserving memory isolation; task conflict detection and priority negotiation mechanisms, enabling multiple application proxy agents to autonomously negotiate and reach consensus in resource competition or goal conflict scenarios; and real-time error correction mechanisms during collaborative execution, enabling the agent network to detect anomalies, diagnose root causes, and dynamically adjust execution strategies during task execution, enhancing the success rate of complex tasks.

**Adaptive Agent Evolution.** The capability boundaries of current application proxy agents are fixed upon initialization. Future work should enable agents to continuously optimize their understanding depth and operational strategies toward applications based on actual usage feedback. This direction explores an adaptive evolution mechanism of "growing stronger through use", including automatic extraction of operational patterns from success and failure cases, identification of application function additions and changes, and optimization of task execution paths based on reinforcement learning, achieving dynamic growth of agent capabilities and synchronous adaptation to application evolution. Voyager achieves lifelong learning for embodied agents in Minecraft environments through automatic curriculum learning and an extensible skill library [52]; the perpetual capability acquisition methodology therein offers significant methodological guidance for the autonomous evolutionary development of application agents.

**Security and Compliance Framework.** Although the current framework has established basic permission isolation and secondary confirmation mechanisms, an end-to-end security governance system has yet to be formed. Future work should construct a comprehensive security framework encompassing agent behavior auditing, operation rollback, data desensitization, and compliance verification, ensuring that the agent network adheres to compliance mandates imposed by regulated sectors including financial services, healthcare provision, and governmental operations, providing trustworthy assurance for large-scale enterprise deployment.

**Large-Scale Deployment Optimization.** The experimental validation of the current framework concentrates on small-to-medium-scale application clusters. Future work should optimize the system architecture for enterprise-level deployment scenarios involving hundreds to thousands of applications. Specific aspects include hierarchical scheduling efficiency improvement for agent networks, dynamic management and elastic scaling of computational resources, together with resilience mechanisms and systematic failure remediation protocols within expansive distributed computational environments, ensuring the stability and scalability of Application-as-Agent in ultra-large-scale heterogeneous system environments.

## 8. Conclusion

This paper introduces Agentic Nesting, an innovative multi-agent collaborative architecture predicated on systematic application abstraction, offering a novel methodological paradigm for the field of enterprise application integration. The term Agentic Nesting denotes the foundational structural principle of our approach: each existing enterprise application is encapsulated as an autonomous AI agent, an operational unit we designate as Application-as-Agent, and multiple such agents are composed within hierarchically nested stewardship layers that mirror the inherent complexity of enterprise ecosystems. The core contributions manifest at three progressive levels.

At the extraction level, we propose a standardized workflow for Application Agent Initialization (AAI), capable of automatically extracting fully functional AI agents from any existing enterprise application, enabling legacy systems originally devoid of intelligent interaction capabilities to acquire natural language dialogue and autonomous manipulation capabilities, achieving automated semantic mapping from application ontology to agent ontology.

At the composition level, this paper formalizes the Agentic Nesting paradigm, designing a hierarchical multi-agent collaborative network architecture where multiple application agents are organized into nested three-layer topologies of "central coordination — application proxy — foundational capability", achieving efficient cross-application collaboration and knowledge sharing through the coordinating agent's intent parsing, task decomposition, and dynamic dispatch, as well as natural language dialogue negotiation among application proxy agents. This systematic architectural characterization elevates the framework from descriptive narrative to systematically structured design, establishing hierarchical coverage as a governing principle for hierarchical agent composition.

At the interaction level, we construct a unified natural language interface that provides users with a single conversational entry point. This shields users from the technical complexity of underlying multi-system environments, enabling cross-application data querying, process orchestration, and intelligent decision-making through natural language descriptions of business intentions. The resulting interaction experience realizes our design philosophy of "Conversation-as-Integration".

Through actual deployment in cross-industry multi-application analysis scenarios, this paper demonstrates the practical applicability of the Agentic Nesting framework in typical scenarios including multi-source data aggregation, cross-system process automation, and heterogeneous data unified querying. More importantly, the methodology of Application-as-Agent possesses broad domain transfer potential, from enterprise software systems to smart home ecosystems, from database management systems to IoT device networks.